\documentclass[runningheads]{llncs}
\usepackage{graphicx}
\usepackage{float}
\usepackage[caption = false]{subfig}

\usepackage{tikz}
\usepackage{comment}
\usepackage{amsmath,amssymb} 
\usepackage{color}
\usepackage{wrapfig}
\usepackage{array}
\usepackage{placeins}
\usepackage{booktabs}

\begin{document}
\pagestyle{headings}
\mainmatter
\def\ECCVSubNumber{7808}  

\title{RRSR:Reciprocal Reference-based Image Super-Resolution with Progressive Feature Alignment and Selection} 

\titlerunning{RRSR}
%
\author{Lin Zhang\thanks{Work done during an internship at Baidu Inc.}\inst{1,3,5,6,*} \and
Xin Li\inst{2,*} \and Dongliang He\inst{2,\dag} \and Fu Li\inst{2} \and Yili Wang\inst{4} \and Zhaoxiang Zhang\inst{1,3,5,7,\dag}
}
\authorrunning{L. Zhang, X. Li et al.}
%
\institute{
$^\text{1 } $ Institute of Automation, Chinese Academy of Sciences \\
$^\text{2 } $  Department of Computer Vision Technology (VIS), Baidu Inc. \\
$^\text{3 } $  University of Chinese Academy of Sciences \qquad $^\text{4 } $ Tsinghua University \\
$^\text{5 } $  National Laboratory of Pattern Recognition, CASIA $^\text{6 } $  School of Future Technology, UCAS \qquad $^\text{7 } $ Center for Artificial Intelligence and Robotics, HKISI\_CAS \\
\email{\{zhanglin2019, zhaoxiang.zhang\}@ia.ac.cn, \{lixin41,hedongliang01,lifu\}@baidu.com, wangyili20@mails.tsinghua.edu.cn} \\
\inst{*}Joint First Authors,~~~\inst{\dag}Joint Corresponding Author
}
\maketitle

\begin{abstract}

Reference-based image super-resolution (RefSR) is a promising SR branch and has shown great potential in overcoming the limitations of single image super-resolution.
While previous state-of-the-art RefSR methods mainly focus on improving the efficacy and robustness of reference feature transfer, it is generally overlooked that a well reconstructed SR image should enable better SR reconstruction for its similar LR images when it is referred to as. Therefore, in this work, we propose a reciprocal learning framework that can appropriately leverage such a fact to reinforce the learning of a RefSR network. 
Besides, we deliberately design a progressive feature alignment and selection module for further improving the RefSR task. The newly proposed module aligns reference-input images at multi-scale feature spaces and performs reference-aware feature selection in a progressive manner, thus more precise reference features can be transferred into the input features and the network capability is enhanced. Our reciprocal learning paradigm is model-agnostic and it can be applied to arbitrary RefSR models. We empirically show that multiple recent state-of-the-art RefSR models can be consistently improved with our reciprocal learning paradigm. Furthermore, our proposed model together with the reciprocal learning strategy sets new state-of-the-art performances on multiple benchmarks. 

\end{abstract}

\section{Introduction}
\label{sec:intro}

\begin{figure}[t]
	\begin{center}
		\includegraphics[width=\columnwidth]{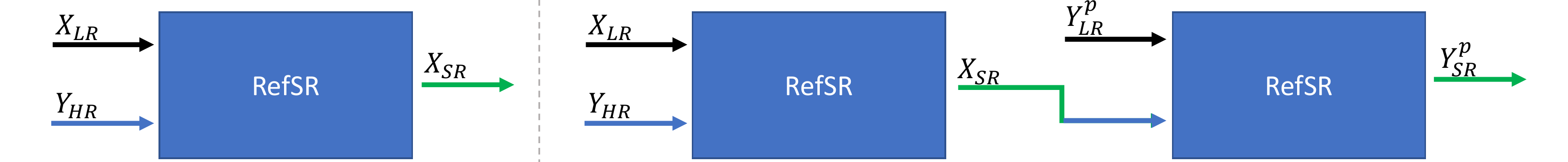}
	\end{center}
	\caption{Left: Traditional RefSR. Right: Our proposed reciprocal training RefSR.}
	\label{fig:fig1}
\end{figure}

Image super-resolution (SR), which aims to reconstruct the corresponding high-resolution (HR) image with natural and sharp details from a low-resolution (LR) image, is an important image processing technique in computer vision. It has broad applications in surveillance~\cite{zhang2010super}, medical imaging~\cite{greenspan2009super}, and astronomy~\cite{holden2011daostorm}, etc. With the prosperity of convolutional neural networks (CNN)~\cite{he2016deep,simonyan2014vggnet,hu2018squeeze}, numerous CNN-based SR methods~\cite{dong2014learning,ledig2017photo,lim2017enhanced,zhang2018image} are proposed, and considerable improvements have been achieved. However, due to the inevitable information loss of the LR images, the SR results often suffer from blurry textures or unrealistic artifacts. As an relaxation to this ill-posed problem, reference-based image super-resolution (RefSR) aims to super-resolve the input LR image with an external HR reference image, which can be acquired from web-searching, photo albums, private repositories, etc. In this manner, similar textures are transferred from the reference image to provide accurate details for the reconstruction of target HR image.

In recent years, there has been extensive research on RefSR. 
A general pipeline for RefSR is as follows: (1) \textit{Search the correlated content in the reference image.} (2) \textit{Align the matched patterns with the input LR features.}  (3) \textit{Fuse the aligned features from reference image into input LR features and then reconstruct target HR image.} To obtain correspondences between the input image and the reference image, some methods~\cite{zheng2018crossnet} directly make use of optical flow, some~\cite{shim2020robust} leverage deformable convolutional networks, and the others~\cite{zhang2019image} perform dense patch matching. $C^2$-Matching~\cite{jiang2021robust} combined dense patch matching with modulated deformable convolution and achieved state-of-the-art performance. MASA~\cite{lu2021masa} proposed a spatial adaptation module to boost the network robustness when there exists a large disparity in distribution. Prior research works focused on leveraging the reference image to the largest extent to improve reconstruction of the target HR image, while little attention is paid to whether the reconstructed SR result can be leveraged to improve the RefSR itself. 

In this paper, we introduce a novel \textbf{R}eciprocal training strategy for \textbf{R}eference-based \textbf{S}uper-\textbf{R}esolution (RRSR) paradigm, as shown in Fig.~\ref{fig:fig1}. Intuitively, if an SR output $X_{SR}$ can be used as reference in turn to boost the performance of super-resolving its similar LR images, $X_{SR}$ should be with clear and sharp context.
Therefore, unlike the previous RefSR methods, we treat the SR output of a RefSR model as reference image and require it to assist in super-resolving a LR variant of the original reference. With such a learning paradigm, the RefSR model can be reinforced to be more robust. 
Besides the reciprocal learning framework, we also propose a \textbf{F}eature \textbf{A}lignment and \textbf{S}election (FAS) module for more accurate reference feature transfer to enhance the capability of the RefSR model. We progressively refine the reference feature alignment at different feature scales by stacking FAS multiple times. Moreover, a set of reference-aware learnable filters is used in FAS for learning to select the most relevant reference features. Our model achieves state-of-the-art performance for the RefSR task. In summary, our contributions are as follows.

\begin{itemize}
    \item To the best of our knowledge, we are the first to introduce reciprocal learning training strategy to RefSR task. We try this reciprocal learning training strategy on multiple RefSR frameworks and achieve consistent performance improvements. 
    
    \item We propose a novel Feature Alignment and Selection (FAS) module for better reference feature utilization. More specifically, we use multiple progressive feature alignment and feature selection with reference-aware learnable filters.
    
    \item Without any bells and whistles, experiments and user studies show that our method obtains favorable performance on several benchmarks. Specially, on the most widely used CUFED5~\cite{wang2016event} dataset, 0.59 dB PSNR gain is achieved compared to prior state-of-the-art.

\end{itemize}

\section{Related Work}
\label{sec:related_work}

In this section, we first briefly introduce the current mainstream research methods for single image super-resolution and reference-based image super-resolution. Then, we discuss weight generating networks and reciprocal learning which are related to our work.

\noindent\textbf{Single Image Super-Resolution.}
In recent years, learning-based approaches with deep convolutional networks achieve promising results in SISR. Dong {\textit{et al}.}~\cite{dong2014learning} first introduced a 3-layer convolutional network to represent the image mapping function between LR images and HR images. Ledig {\textit{et al}.}~\cite{ledig2017photo} used residual blocks which are originally designed for high-level tasks and brought a significant reduction in reconstruction error. With elaborate analysis, lim {\textit{et al}.}~\cite{lim2017enhanced} removed the batch normalization layers~\cite{ioffe2015batch} and developed a new training procedure to achieve better SR performance. Zhang {\textit{et al}.}~\cite{zhang2018image} and Dai {\textit{et al}.}~\cite{dai2019second} introduced channel attention~\cite{hu2018squeeze} to explore inter-channel correlations. Recently, 
a lot of works~\cite{liu2018non,dai2019second,zhang2019residual,mei2020image,mei2021image} adopted non-local attention to model long-range feature relationships, further improving SR performance. Moreover, the PSNR-oriented methods lead to overly-smooth textures, another branch of works aiming at improving the perceptual quality have been proposed. Johnson {\textit{et al}.}~\cite{johnson2016perceptual} combined MSE with the perception loss based on high-level convolutional features~\cite{simonyan2014vggnet}. Generative adversarial network (GAN)~\cite{goodfellow2014gan} prior was also introduced into SR tasks by~\cite{ledig2017photo} and further refined by~\cite{sajjadi2017enhancenet,wang2018esrgan,zhang2019ranksrgan}.

\noindent\textbf{Reference-based Image Super-Resolution.}
Different from SISR, whose only input is an LR image, RefSR~\cite{yue2013landmark,zheng2018crossnet} use an additional reference image, which greatly improves the SR reconstruction.
The RefSR methods transfer the fine details of the external reference images to the regions with similar textures in the input LR image, so that the SR reconstruction obtains more abundant high-frequency components.
Zhang {\textit{et al}.}~\cite{zhang2019image} performed a multi-scale feature transfer by conducting local patch matching in the feature space and fusing multiple swapped features to the input LR features. This enables the network with strong robustness even when irrelevant reference images are given. Subsequently, TTSR~\cite{yang2020learning} unfroze the parameters of VGG extractor~\cite{simonyan2014vggnet} and proposed a cross-scale feature integration module to merge multi-scale features, which enhances the feature representation. MASA~\cite{lu2021masa} designed a coarse-to-fine patch matching scheme to reduce the computational complexity and a spatial adaption module to boost the robustness of the network when the input and the reference differ in color distributions.

$C^2$-Matching~\cite{jiang2021robust} proposed a contrastive correspondence network to perform scale and rotation robust matching between input images and reference images. Although $C^2$-Matching greatly improves the matching accuracy, there is still room for improvement in terms of alignment. Besides, $C^2$-Matching did not filter and select the reference features and ignored the potential large disparity in distributions between the LR and Ref images. We conduct a multiple times progressive tuning at each feature scale to further improve reference feature alignment. Besides, we design some reference-aware learnable filters to select reference features.

\noindent\textbf{Weight Generating Networks}
Unlike classical neural networks, where the weight parameters are frozen after training, weight generating networks~\cite{ha2016hypernetworks,jia2016dynamic,ma2020weightnet,yang2019condconv} dynamically produce the weight parameters conditioned on the latent vectors or the input. Ha {\textit{et al}.}~\cite{ha2016hypernetworks} proposed the HyperNetworks that uses an extra network to generate the weight parameters for the main network.
Instead of learning input-agnostic weights, Jia {\textit{et al}.}~\cite{jia2016dynamic} and Yang {\textit{et al}.}~\cite{yang2019condconv} suggested learning different weights for different samples.
Thanks to its powerful representational capability and customizability, weight generating networks have been modified successfully for image denoising~\cite{mildenhall2018burst,xia2020basis}, instance segmentation~\cite{tian2020conditional}, neural rendering~\cite{sitzmann2019scene} and scale-arbitrary SR~\cite{hu2019meta,wang2021learning}. 
In this work, we extend the idea to generate a set of reference-aware filters for reference feature selection.

\noindent\textbf{Reciprocal Learning.} Unlike these RefSR methods which focused on reference feature transfer, we propose a reciprocal training strategy by using a RefSR result image as a new reference and conducting a RefSR for the second time. Throughout the literature of reciprocal learning~\cite{he2016dual,sun2020reciprocal,jiang2021reciprocal,pham2021meta,zagalsky2021design}, it generally involves a pair of parallel learning processes, excavates the relation between them and constructs a learning-feedback loop, thus promoting the primal learning process. 
In neural machine translation, He {\textit{et al}.}~\cite{he2016dual} created a closed-loop, learning English-to-French translation (source $\to$ target) versus learning French-to-English translation (target $\to$ source), and both learning processes are improved by generated feedback signals. Similar strategies have been applied successfully to unpaired image-to-image translation such as DualGAN~\cite{yi2017dualgan}. More recently, Sun {\textit{et al}.}~\cite{sun2020reciprocal} designed two coupling networks, one predicting future human trajectories and the other for past human trajectories, and achieved great improvement on human trajectory prediction task by jointly optimizing two models under the reciprocal constraint. In this paper, we make a conjecture that an ideal SR result can also served as a new reference image and provide high-frequency information to other similar LR images. Hence, we propose a reciprocal training strategy for RefSR.

\begin{figure}[t]
	\begin{center}
		\includegraphics[width=\columnwidth]{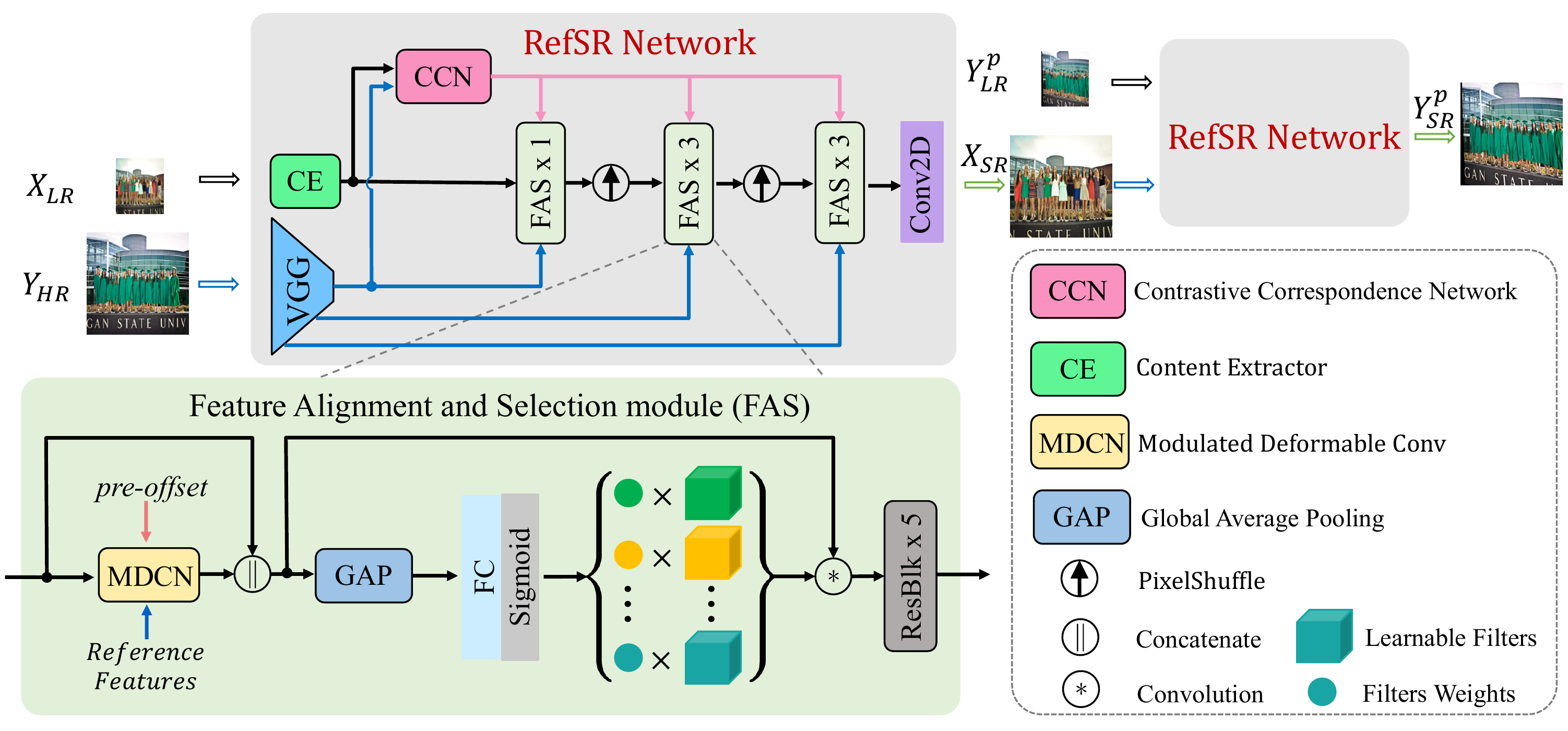}
	\end{center}
	\caption{An overview of RRSR with \textbf{Reciprocal Target-Reference Reconstruction} (top) and \textbf{Progressive Feature Alignment and Selection} (bottom-left). By creating an extra branch that super-resolves reference LR, Reciprocal Target-Reference Reconstruction constructs a dual task, i.e., reference $\to$ target and target $\to$ reference, thus improving both target reconstruction and reference reconstruction. We progressively refine the reference feature alignment at $2\times$ and $4\times$ feature scales by stacking FAS multiple times. Moreover, a set of reference-aware learnable filters is used in FAS for learning to select the most relevant reference features.}
	\label{fig:Method}
\end{figure}
\section{Approach}
\label{sec:approach}

The overall architecture of the proposed method is shown in Fig.~\ref{fig:Method}. Reciprocal learning paradigm is designed to boost the training of the reference-based image super-resolution network (RefSR Network). Intuitively, if a super-resolved output is well reconstructed, it can be qualified to serve as reference image for super-resolving the low-resolution version of the original reference image. Therefore, in our framework, the original high-resolution reference image $Y_{HR}$ is leveraged to help the reconstruction of high-resolution version $X_{SR}$ for a given low-resolution input $X_{LR}$. In turn, we treat $X_{SR}$ as reference and require the network to well reconstruct $Y_{SR}^{p}$ by super-resolving its low-resolution counterpart $Y_{LR}^{p}$. Note that, we did not directly super-resolve $Y_{LR}$, but a warped version $Y_{LR}^p$ is fed into the RefSR network. Such a learning paradigm is termed as \textbf{R}eciprocal \textbf{T}arget-\textbf{R}eference \textbf{R}econstruction (RTRR) in this paper. 

Our RefSR network is largely inspired by the $C^2$-Matching~\cite{jiang2021robust}, meanwhile, a novel feature alignment and selection module (FAS) is proposed and progressively stacked for more accurate reference feature transformation. Specifically, a VGG network and a context encoder (CE) are used to encode the high-resolution reference image and the input low-resolution image, respectively. Then, a contrastive correspondence network (CCN) is applied for predicting the pre-offset of input and reference at feature space. Subsequently, our proposed FAS module takes as input the pre-offset, input image feature, and the reference feature at different scales to progressively transfer reference features to the input feature space for high-resolution output reconstruction. 
We will present details of our reciprocal learning paradigm and the feature alignment and selection module in Sec.~\ref{sec:reciprocal_reconstrution} and Sec.~\ref{sec:feature_transfer}, respectively.

\subsection{Reciprocal Target-Reference Reconstruction}
\label{sec:reciprocal_reconstrution}
Given the success of reciprocal learning in many research fields, we postulate that RefSR would also benefit from a carefully designed reciprocal learning strategy since the roles of input and reference could switch mutually. Intuitively, if a super-resolved image is well reconstructed with clear and sharp context, it should be suitable to serve as reference image to guide the super-resolution reconstruction of other similar LR images. Specifically, for reference-based image super-resolution scenario, a straightforward way to compose a reciprocal learning framework is to use the output image $X_{SR}$, which is reconstructed by referring to the original reference image $Y_{HR}$ via a RefSR network, as reference for super-resolving the down-sampled reference image $Y_{LR}$ by using the RefSR a second time to generate $Y_{SR}$. We adopt $\ell_1$ loss as the reconstruction objective, then the two pass RefSR reconstruction and reciprocal loss to be optimized can be represented as:

\begin{equation}
\mathcal{L}_{rec} = \|X_{HR} - X_{SR}\|_{1}, ~~~~X_{SR} = RefSR(X_{LR}, Y_{HR}),
\label{eq.1}
\end{equation}

\begin{equation}
\mathcal{L}_{RTRR} = \|Y_{HR} - Y_{SR}\|_{1}, ~~~~Y_{SR} = RefSR(Y_{LR}, X_{SR}),
\label{eq.2}
\end{equation}
Note that, during the two phases, the parameters of RefSR networks are shared.

However, with such straightforward configuration, the reciprocal learning framework will collapse. That is because $Y_{HR}$ is the input of the first RefSR stage as reference, then the whole process combining Eq.\ref{eq.1} and Eq.\ref{eq.2} becomes an auto-encoder for $Y_{HR}$. The auto-encoder will push the first RefSR stage to keep $X_{SR}$ having as much $Y_{HR}$ information as possible for better reconstructing $Y_{SR}$, and the $L_{RTRR}$ drops quickly such that the capability of RefSR network is not enhanced. 

\begin{figure}[t]
	\begin{center}
		\includegraphics[width=\columnwidth]{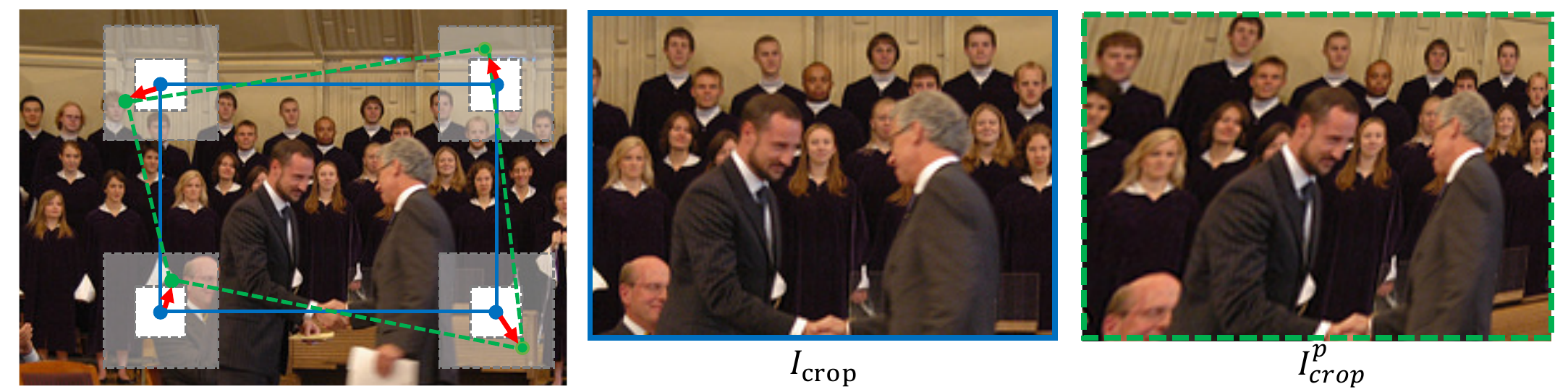}
	\end{center}
	\caption{Illustration for perspective transformation. Given an image $I$ and {\color{blue}a rectangular bounding box}, we can get {\color{green}four new vertices} by randomly perturbing the four vertices of the box in {\color{gray}a fixed area}. Then we can use these two sets of vertices to compute an perspective transformation matrix which is used to transfer $I$ to $I^p$. Finally we crop $I^p$ with the original box to get a perspective transformation variant $I^p_{crop}$ of $I_{crop}$.}
	\label{fig:pt}
\end{figure}

To prevent RefSR from collapsing, we introduce a simple yet effective mechanism for processing $Y_{HR}$. In the second RefSR phase, under the condition that $X_{SR}$ is the reference image, we apply a random perspective transformation on the original $Y_{LR}$ and $Y_{HR}$ to obtain the training image pair ($Y_{LR}^{\!\mathcal{P}}$, $Y_{HR}^{\!\mathcal{P}}$). A process of perspective transformation is shown in Fig.~\ref{fig:pt}. Because $Y_{SR}^{\!\mathcal{P}}$ and $Y_{HR}$ are quite different, the reconstruction of $Y_{SR}^{\!\mathcal{P}}$ will not force $X_{SR}$ to be same as $Y_{HR}$. The revised RTRR loss then becomes as follows:

\begin{equation}
\label{equa:revised_loss}
\mathcal{L}_{RTRR} = \|Y_{HR}^{\!\mathcal{P}} - Y_{SR}^{\!\mathcal{P}}\|_{1}, ~~~~Y_{SR}^{\!\mathcal{P}} = RefSR(Y_{LR}^{\!\mathcal{P}}, X_{SR}).
\end{equation}

To minimize $\mathcal{L}_{RTRR}$, $Y_{SR}^{\!\mathcal{P}}$ should be well reconstructed, which in turn depends on the quality of its reference image $X_{SR}$. 
Thus optimizing the RefSR result $Y_{SR}^{\!\mathcal{P}}$ at the second phase can also help optimize the first SR result $X_{SR}$. 
The proposed reciprocal target-reference reconstruction (RTRR) training strategy is model-agnostic and can be leveraged to improve arbitrary reference-based SR networks. We empirically show improvements of $C^2$-Matching~\cite{jiang2021robust}, MASA-SR~\cite{lu2021masa} and TTSR~\cite{yang2020learning} by using our RTRR configuration in Sec.~\ref{sec:ablation}.

\subsection{Progressive Feature Alignment and Selection}
\label{sec:feature_transfer}
Our RefSR network is based on $C^2$-Matching \cite{jiang2021robust} considering its state-of-the-art performance. We also propose progressive feature alignment and selection to further improve its capability. As shown in Fig.~\ref{fig:Method}, a \emph{Content Extractor} is used to extract features $F_{X_{LR}}$ from $X_{LR}$. Multi-scale ($1\times$, $2\times$ and $4\times$) reference image features $F^s_{Y_{HR}}$ are extracted by a \emph{VGG} extractor, where $s = 1, 2, 4$. A pretrained \emph{Contrastive Correspondence Network} is used to obtain the relative target offsets of the LR input and reference images. These offsets are used as \emph{pre-offsets} for reference feature alignment. We use the \emph{Modulated Deformable Convolution} (MDCN) in $C^2$-Matching~\cite{jiang2021robust} to align the reference features. But unlike $C^2$-Matching~\cite{jiang2021robust}, we propose a progressive feature alignment and selection module (PFAS) for better aggregating information from the reference image.

First, an MDCN is used to initially align reference features $F^s_{Y_{HR}}$ to LR image features by the \emph{pre-offsets}. Then the aligned reference features and the LR features are concatenated together to get a merged feature $F_{merge}^s$.
In real scenarios, there may be many differences between input and reference images in terms of color, style, intensity, etc. Therefore, the features of the reference images should be elaborately selected. 
We design a reference-aware feature selection mechanism to selectively exploit reference features. Reference-aware information is obtained by applying a global average pooling $GAP$ to $F_{merge}^s$. Then it is used to generate routing weights:
\begin{equation}
\label{equa:routing}
(\alpha_1, \alpha_2, \ldots, \alpha_K) = \sigma(f(GAP(F_{merge}^s))) \,,
\end{equation}
where $f(\cdot)$ and $\sigma(\cdot)$ denote fully-connected layer and sigmoid activation. These routing weights are expected to combine $K$ learnable template filters $E_k, k\in\{1,2,...,K\}$ which are applied onto $F_{merge}$ to choose the features:
\begin{equation}
\label{equa:select_feature}
\begin{aligned}
F_{selected}^s &= \alpha_1(E_1 \ast F_{merge}^s) + \alpha_2(E_2 \ast F_{merge}^s) + \ldots + \alpha_K(E_K \ast F_{merge}^s) \\
             &= (\alpha_1 E_1 + \alpha_2 E_2 + \ldots + \alpha_K E_K) \ast F_{merge}^s\,,
\end{aligned}             
\end{equation}
where $\ast$ denotes convolution operation.
It can be seen that it is equivalent to predict a reference-aware filter, which is eventually used to process the $F_{merge}^s$ for feature selection. Unlike static convolution, the reference-aware filter is adaptively conditioned on input and reference. It is demonstrated in Sec.~\ref{sec:ablation} that our network benefits from the reference-aware feature selection to produce better results for reference-based SR. At the last of the module, there are several residual blocks to fuse the selected and aligned reference features with LR features. 
Furthermore, we use the module three times at the $2\times$ and $4\times$ scale to progressively refine the feature alignment and selection. In this way, the details in the LR and reference features are enhanced and aggregated.

\subsection{Loss Functions}
\label{sec:loss}

Our overall objective is formulated as
\begin{equation}
\label{equa:all_loss}
\mathcal{L} = \lambda_{rec}\mathcal{L}_{rec} + \lambda_{RTRR}\mathcal{L}_{RTRR} + \lambda_{per}\mathcal{L}_{per} + \lambda_{adv}\mathcal{L}_{adv}\,.
\end{equation}

The reconstruction loss $\mathcal{L}_{rec}$ and the reciprocal loss $\mathcal{L}_{RTRR}$ have been introduced in Sec.~\ref{sec:reciprocal_reconstrution}. To generate sharp and visually pleasing images, we employ perceptual loss $\mathcal{L}_{per}$ and adversarial loss $\mathcal{L}_{adv}$ introduced in~\cite{zhang2019image} to help the training of the network.

\section{Experiments}
\label{sec:experiments}

\subsection{Datasets and Metrics}
Following~\cite{zhang2019image,yang2020learning,shim2020robust,jiang2021robust}, we use the training set of CUFED5~\cite{zhang2019image,wang2016event} as our training dataset, which contains 11,781 training pairs. Each pair consists of a target HR image and a corresponding reference HR image, both at about 160x160 resolution. We evaluate SR results on the testing set of CUFED5, Sun80~\cite{sun2012super}, Urban100~\cite{huang2015single}, Manga109~\cite{matsui2017sketch} and the newly proposed WR-SR~\cite{jiang2021robust}. 
The testing set of CUFED5 contains 126 pairs, and each consists of an HR image and five reference images with different similarity levels based on SIFT feature matching. In order to be consistent with the previous methods, we pad each reference image to the left-top corner in each 500x500 zero image and stitch them to obtain a 2500x500 image as the reference image. The Sun80 dataset contains 80 natural images, and each with 20 web-searching references. The reference image is randomly selected from them. The Urban100 dataset contains 100 building images, lacking references. Because of self-similarity in the building image, the corresponding LR image is treated as the reference image.
The Manga109 dataset contains 109 manga images without references. Since all the images in Manga109 are the same category (manga cover) and some similar patterns occur across the images, we randomly sample an HR image from the dataset as the reference image. The WR-SR dataset, which is proposed by~\cite{jiang2021robust} to cover more diverse categories, contains 80 image pairs, each target image accompanied by a web-searching reference image. All the LR images are obtained by bicubically downsampling the HR images with the scale factor $4\times$.
All the results are evaluated in PSNR and SSIM on Y channel in the transformed YCbCr color space.

\begin{table}[!t]
\centering
\caption{We report PSNR/SSIM (higher is better) of different SR
methods on the testing set of CUFED5~\cite{zhang2019image,wang2016event}, Sun80~\cite{sun2012super}, Urban100~\cite{huang2015single}, Manga109~\cite{matsui2017sketch}, and WR-SR~\cite{jiang2021robust}. Methods are grouped by SISR methods (top) and reference-based methods (bottom). 
Urban100 indicated with $\dagger$ lacks an external reference image and all the methods essentially degrade to be SISR methods. The best and the second best results are shown in {\color{red}red} and {\color{blue}blue}, respectively.}
\label{table:results}  
\resizebox{\textwidth}{!}{%

\begin{tabular}{l|c|c|c|c|c}
\toprule
& CUFED5~\cite{zhang2019image,wang2016event} & Sun80~\cite{sun2012super} & Urban100$\dagger$~\cite{huang2015single} & Manga109~\cite{matsui2017sketch} & WR-SR~\cite{jiang2021robust} \\
Method & PSNR$\uparrow$ / SSIM$\uparrow$ & PSNR$\uparrow$ / SSIM$\uparrow$ & PSNR$\uparrow$ / SSIM$\uparrow$ & PSNR$\uparrow$ / SSIM$\uparrow$ & PSNR$\uparrow$ / SSIM$\uparrow$ \\
\hline
SRCNN~\cite{dong2014learning} & $25.33$ / $0.745$ & $28.26$ / $0.781$ & $24.41$ / $0.738$ & $27.12$ / $0.850$ & $27.27$ / $0.767$ \\
EDSR~\cite{lim2017enhanced} & $25.93$ / $0.777$ & $28.52$ / $0.792$ & $25.51$ / $0.783$ & $28.93$ / $0.891$ & $28.07$ / $0.793$ \\
RCAN~\cite{zhang2018image} & $26.06$ / $0.769$ & $29.86$ / $0.810$ & $25.42$ / $0.768$ & $29.38$ / $0.895$ & $28.25$ / $0.799$ \\
NLSN~\cite{mei2021image} & $26.53$ / $0.784$ & {\color{blue}$30.16$} / {\color{blue}$0.816$} & {\color{red}$26.28$} / {\color{red}$0.793$} & {\color{blue}$30.47$} / {\color{blue}$0.911$} & $28.07$ / $0.794$ \\
SRGAN~\cite{ledig2017photo} & $24.40$ / $0.702$ & $26.76$ / $0.725$ & $24.07$ / $0.729$ & $25.12$ / $0.802$ & $26.21$ / $0.728$ \\
ENet~\cite{sajjadi2017enhancenet} & $24.24$ / $0.695$ & $26.24$ / $0.702$ & $23.63$ / $0.711$ & $25.25$ / $0.802$ & $25.47$ / $0.699$ \\
ESRGAN~\cite{wang2018esrgan} & $21.90$ / $0.633$ & $24.18$ / $0.651$ & $20.91$ / $0.620$ & $23.53$ / $0.797$ & $26.07$ / $0.726$ \\
RankSRGAN~\cite{zhang2019ranksrgan} & $22.31$ / $0.635$ & $25.60$ / $0.667$ & $21.47$ / $0.624$ & $25.04$ / $0.803$ & $26.15$ / $0.719$ \\
\hline
CrossNet~\cite{zheng2018crossnet} & $25.48$ / $0.764$ & $28.52$ / $0.793$ & $25.11$ / $0.764$ & $23.36$ / $0.741$ & - \\
SRNTT~\cite{zhang2019image} & $25.61$ / $0.764$ & $27.59$ / $0.756$ & $25.09$ / $0.774$ & $27.54$ / $0.862$ & $26.53$ / $0.745$ \\
SRNTT-$rec$~\cite{zhang2019image} & $26.24$ / $0.784$ & $28.54$ / $0.793$ & $25.50$ / $0.783$ & $28.95$ / $0.885$ & $27.59$ / $0.780$ \\
TTSR~\cite{yang2020learning} & $25.53$ / $0.765$ & $28.59$ / $0.774$ & $24.62$ / $0.747$ & $28.70$ / $0.886$ & $26.83$ / $0.762$ \\
TTSR-$rec$~\cite{yang2020learning} & $27.09$ / $0.804$ & $30.02$ / $0.814$ & $25.87$ / $0.784$ & $30.09$ / $0.907$ & $27.97$ / $0.792$ \\
$C^2$-Matching~\cite{jiang2021robust} & $27.16$ / $0.805$ & $29.75$ / $0.799$ & $25.52$ / $0.764$ & $29.73$ / $0.893$ & $27.80$ / $0.780$ \\
$C^2$-Matching-$rec$~\cite{jiang2021robust} & {\color{blue}$28.24$} / {\color{blue}$0.841$} & {\color{red}$30.18$} / {\color{red}$0.817$} & $26.03$ / $0.785$ & {\color{blue}$30.47$} / {\color{blue}$0.911$} & {\color{blue}$28.32$} / {\color{blue}$0.801$} \\
MASA~\cite{lu2021masa} & $24.92$ / $0.729$ & $27.12$ / $0.708$ & $23.78$ / $0.712$ & $27.44$ / $0.849$ & $25.76$ / $0.717$ \\
MASA-$rec$~\cite{lu2021masa} & $27.54$ / $0.814$ & $30.15$ / $0.815$ & $26.09$ / $0.786$ & $30.28$ / $0.909$ & $28.19$ / $0.796$ \\
SSEN~\cite{shim2020robust} & $25.35$ / $0.742$ & - & - & - & - \\
SSEN-$rec$~\cite{shim2020robust} & $26.78$ / $0.791$ & - & - & - & - \\
DCSR~\cite{wang2021dual} & $25.39$ / $0.733$ & - & - & - & - \\
DCSR-$rec$~\cite{wang2021dual} & $27.30$ / $0.807$ & - & - & - & - \\
Ours & $28.09$ / $0.835$ & $29.57$ / $0.793$ & $25.68$ / $0.767$ & $29.82$ / $0.893$ & $27.89$ / $0.784$ \\
Ours-$rec$ & {\color{red}$28.83$} / {\color{red}$0.856$} & $30.13$ / {\color{blue}$0.816$} & {\color{blue}$26.21$} / {\color{blue}$0.790$} & {\color{red}$30.91$} / {\color{red}$0.913$} & {\color{red}$28.41$} / {\color{red}$0.804$} \\
\bottomrule
\end{tabular}
}
\end{table}

\subsection{Training Details}
We train our RefSR network for 255K iterations with a mini-batch size of 9, using Adam optimizer with parameters $\beta_1=0.9$ and $\beta_2=0.999$.  The initial learning rate is set to 1e-4. The weight coefficients for $\lambda_{rec}$, $\lambda_{RTRR}$, $\lambda_{per}$ and $\lambda_{adv}$ are set to 1, 0.4, 1e-4 and 1e-6, respectively. The perspective transformation perturbation range in RTRR of the vertex is [-20, -5] and [5, 20] both horizontally and vertically. The reason why [-5, 5] is not used because the perturbation must be guaranteed to exceed a certain magnitude, otherwise the perturbed image $Y_{LR}^{\!\mathcal{P}}$ is too similar to $Y_{LR}$, which hinders the performance. The number of learnable filters in each FAS module is 16. In order to keep the network complexity close to $C^2$-Matching, the number of res-blocks in a FAS is 5, and other network parameters are the same as $C^2$-Matching. Please refer to supplementary material for the network implementation details. The input LR patch size is 40x40, corresponding to a 160x160 ground-truth HR patch. During training, we augment the training data with randomly horizontal flipping, randomly vertical flipping, and random {90\textdegree} rotation.

\subsection{Comparison with State-of-the-Art Methods}
\label{sec:results}

\begin{figure}[!t]
    \centering
	\begin{tabular}{p{0.222\textwidth}<{\centering}p{0.221\textwidth}<{\centering}p{0.221\textwidth}<{\centering}p{0.221\textwidth}<{\centering}}
		Input LR & ESRGAN~\cite{wang2018esrgan} & RankSRGAN~\cite{zhang2019ranksrgan} & TTSR~\cite{yang2020learning} \\
		\hline
		Reference HR & MASA~\cite{lu2021masa} & $C^2$-Matching~\cite{jiang2021robust} & Ours \\
	\end{tabular}
    \includegraphics[width=0.905\textwidth]{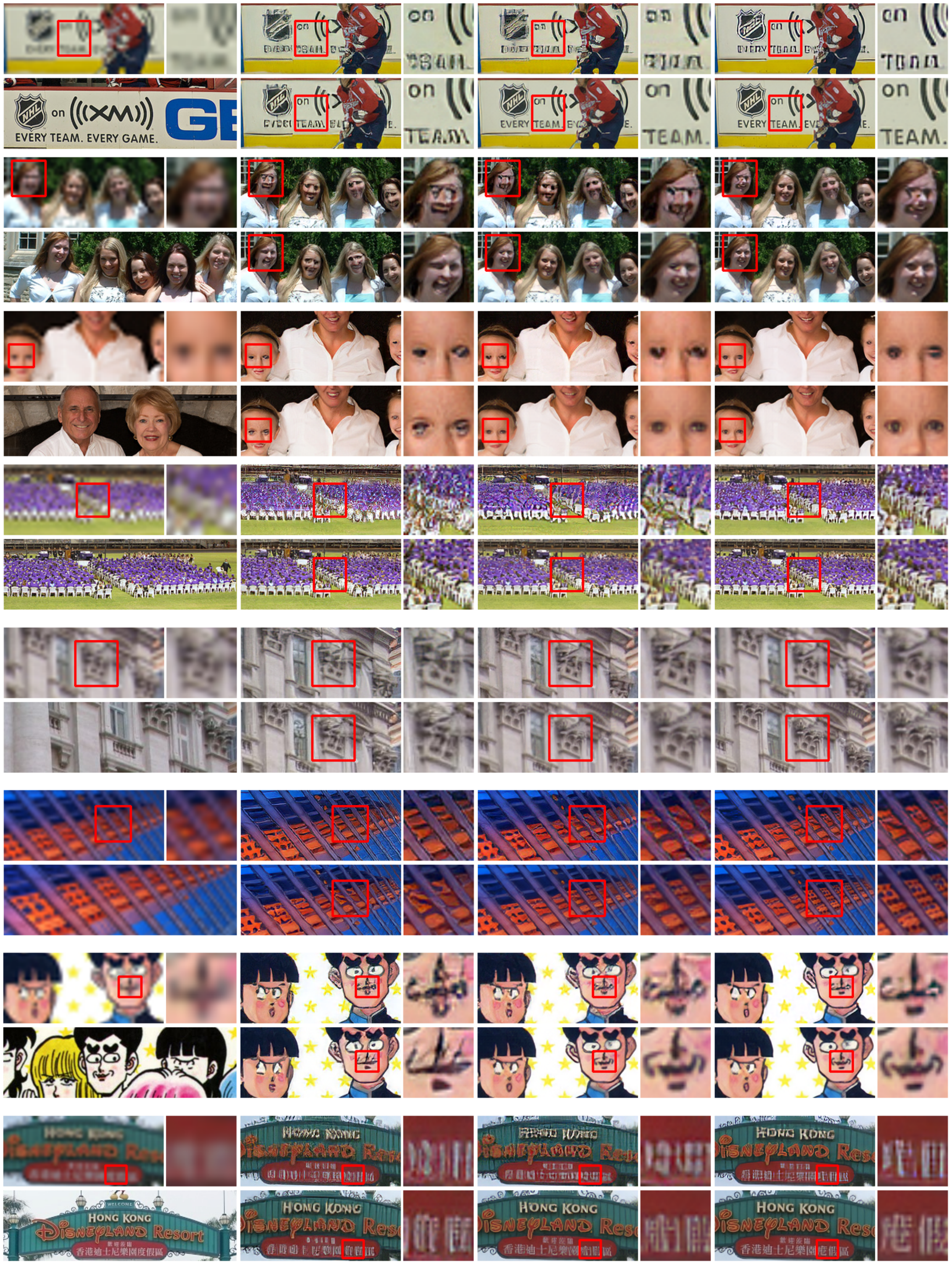}
    \caption{Comparisons on the testing set of CUFED5~\cite{zhang2019image,wang2016event} (the first four examples), Sun80~\cite{sun2012super} (the fifth example), Urban100~\cite{huang2015single} (the sixth example), Manga109~\cite{matsui2017sketch} (the seventh example) and WR-SR~\cite{jiang2021robust} (the eighth example). Our method is able to discover highly related content in reference HR and properly transfer it to restore sharper and more natural textures than the prior works.}

    \label{fig:results}
\end{figure}

\noindent\textbf{Quantitative Comparison.}
We quantitatively compare our method with previous state-of-the-art SISR methods and reference-based SR methods. 
SISR methods include SRCNN~\cite{dong2014learning}, EDSR~\cite{lim2017enhanced}, RCAN~\cite{zhang2018image}, NLSN~\cite{mei2021image},  SRGAN~\cite{ledig2017photo}, ENet~\cite{sajjadi2017enhancenet}, ESRGAN~\cite{wang2018esrgan}, and RankSRGAN~\cite{zhang2019ranksrgan}. As for reference-based SR methods, CrossNet~\cite{zheng2018crossnet}, SRNTT~\cite{zhang2019image}, TTSR~\cite{yang2020learning}, SSEN~\cite{shim2020robust}, DCSR~\cite{wang2021dual}, $C^2$-Matching~\cite{jiang2021robust}, and MASA~\cite{lu2021masa} are included. All the reference-based methods except CrossNet have a PSNR-oriented variant (training without GAN loss and perceptual loss), marked with the suffix '-$rec$'.

As shown in Table~\ref{table:results}, our proposed method can outperform almost all comparative methods. On the standard CUFED5 benchmark, our method shows a significant improvement of 0.59 dB over the previous state-of-the-art $C^2$-Matching. On the Sun80 and Urban100 datasets, our method performs comparably to the state-of-the-art methods. Because on the Urban100 dataset, the reference image is the input image itself, reference-based methods have no advantage over SISR methods. As for the Manga109 dataset, the performance of our method surpasses the second-place candidate by a large margin of 0.44 dB. Moreover, our method still achieves the best performance on the WR-SR dataset.

\noindent\textbf{Qualitative Comparison.}
Fig.~\ref{fig:results} shows some visual results for qualitative comparisons. We compare our method with current top-performing methods, ESRGAN~\cite{wang2018esrgan}, RankSRGAN~\cite{zhang2019ranksrgan}, TTSR~\cite{yang2020learning}, MASA~\cite{lu2021masa}, and $C^2$-Matching~\cite{jiang2021robust}. As demonstrated by the examples, our method can extract the correlated information from reference HR image and correctly transfer it to finer HR reconstruction in both sharpness and details. As shown in the second example, our approach recovers a clear plausible face while other methods fail. Even though $C^2$-Matching achieves a comparable result, artifacts appeared in the left eye. Besides, as shown in the bottom example, only the proposed method recovers the exact two Chinese characters. More visual comparisons are provided in supplementary material.

\begin{figure}[t]
    \centering
    \subfloat[]{\includegraphics[height=4.64cm]{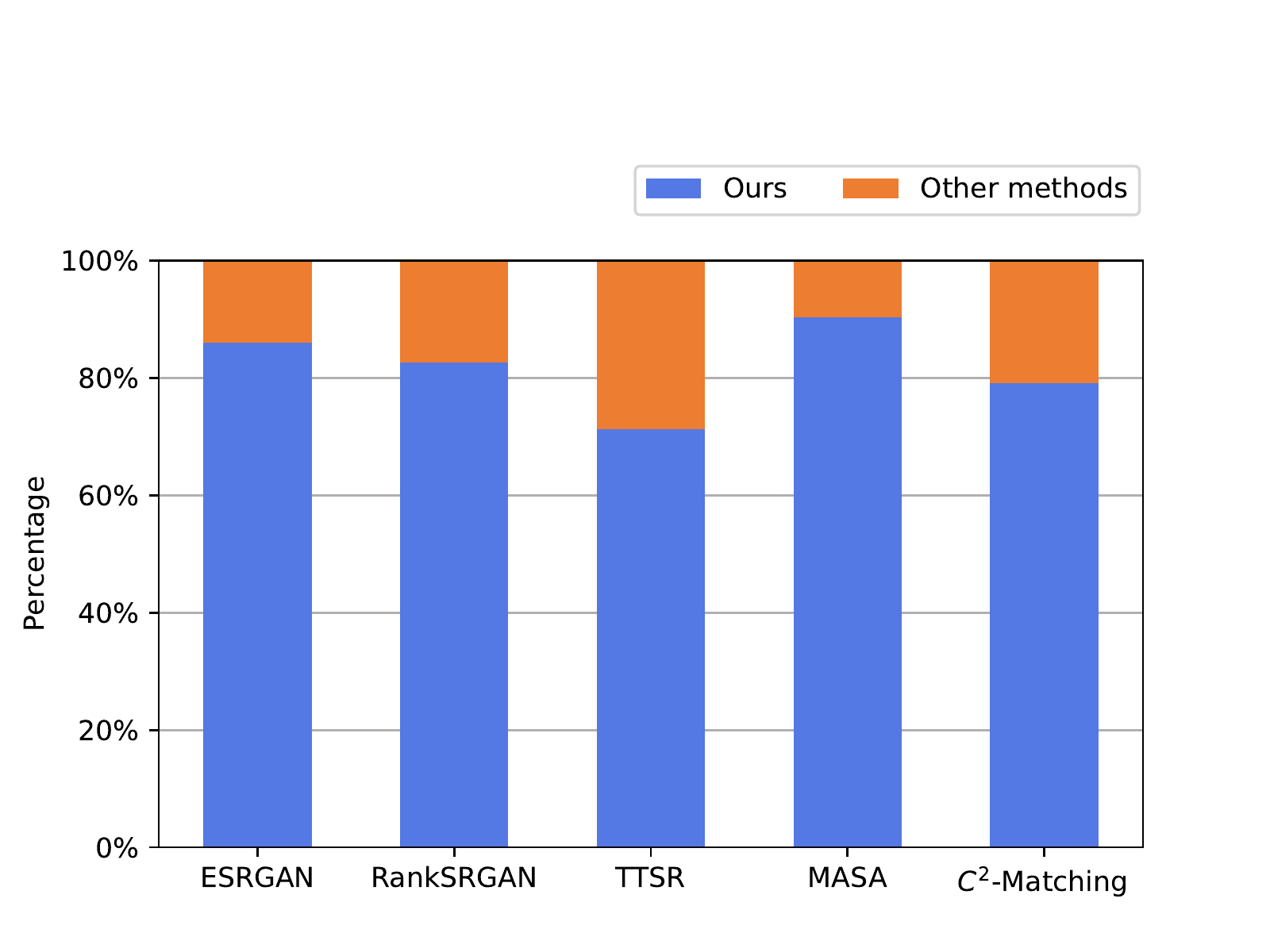}} \,
    \subfloat[]{\includegraphics[height=4.26cm]{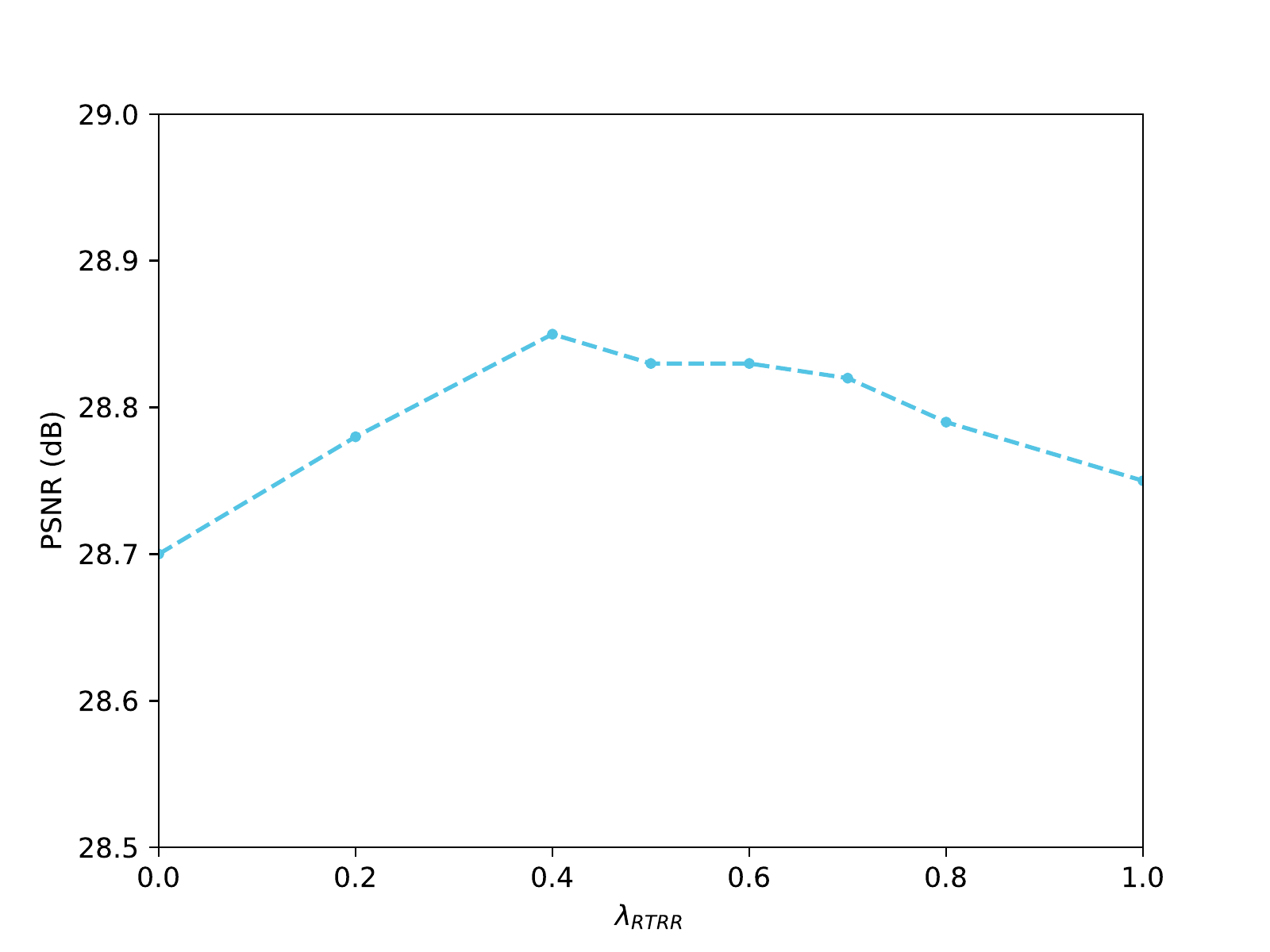}}
\caption{(a) User study results. Values on Y-axis indicate the voting percentage for preferring our method. (b) Influence of $\lambda_{RTRR}$.}
\label{fig:user_loss}
\end{figure}

Following the convention, we also conduct a user study to compare our method with the above methods. Exactly, in each test, we present every participant with two super-resolution results, one predicted by our method and another predicted by one of the other methods. A total number of 50 users are asked to compare the visual quality. As shown in Fig.~\ref{fig:user_loss} (a), the participants favor our results against other state-of-the-arts methods.

\begin{table}[b]
\caption{Left: Ablation study to analyze the effectiveness of each component of our RRSR. Right: Ablation study on our reciprocal target-reference reconstruction. Asterisks represent our achieved results, not official results.}
\label{table:ablations}
\begin{minipage}{0.49\textwidth}
\qquad
\centering
\scalebox{0.77}{
\begin{tabular}{l|c|c|c|c}
\hline
Model &PFA &RAS &RTRR &PSNR $\uparrow$ / SSIM$\uparrow$ \\
\hline
Baseline($C^2$-Matching$^*$) &            &            &            & $28.40$ / $0.846$ \\
Baseline+PFA            & \checkmark &            &            & $28.63$ / $0.851$ \\
Baseline+PFA+RAS        & \checkmark & \checkmark &            & $28.70$ / $0.853$ \\
Baseline+PFA+RAS+RTRR   & \checkmark & \checkmark & \checkmark & $28.83$ / $0.856$ \\
\hline
\end{tabular}
}
\qquad
\end{minipage}
\begin{minipage}{0.03\textwidth}
\qquad
\end{minipage}
\begin{minipage}{0.49\textwidth}
\qquad
\centering
\scalebox{0.63}{
\begin{tabular}{l|c|c}
\hline
Model &RTRR &PSNR $\uparrow$ / SSIM$\uparrow$ \\
\hline
$C^2$-Matching$^*$      &            & $28.40$ / $0.846$ \\ 
$C^2$-Matching$^*$+RTRR & \checkmark & $28.64$ / $0.853$ \\
MASA$^*$             &            & $27.47$ / $0.815$ \\ 
MASA$^*$+RTRR        & \checkmark & $27.58$ / $0.818$ \\
TTSR$^*$             &            & $27.03$ / $0.800$ \\ 
TTSR$^*$+RTRR        & \checkmark & $27.17$ / $0.804$ \\
\hline
\end{tabular}
}
\end{minipage}

\end{table}

\begin{figure}[t]
	\begin{center}
		\includegraphics[height=5.6cm]{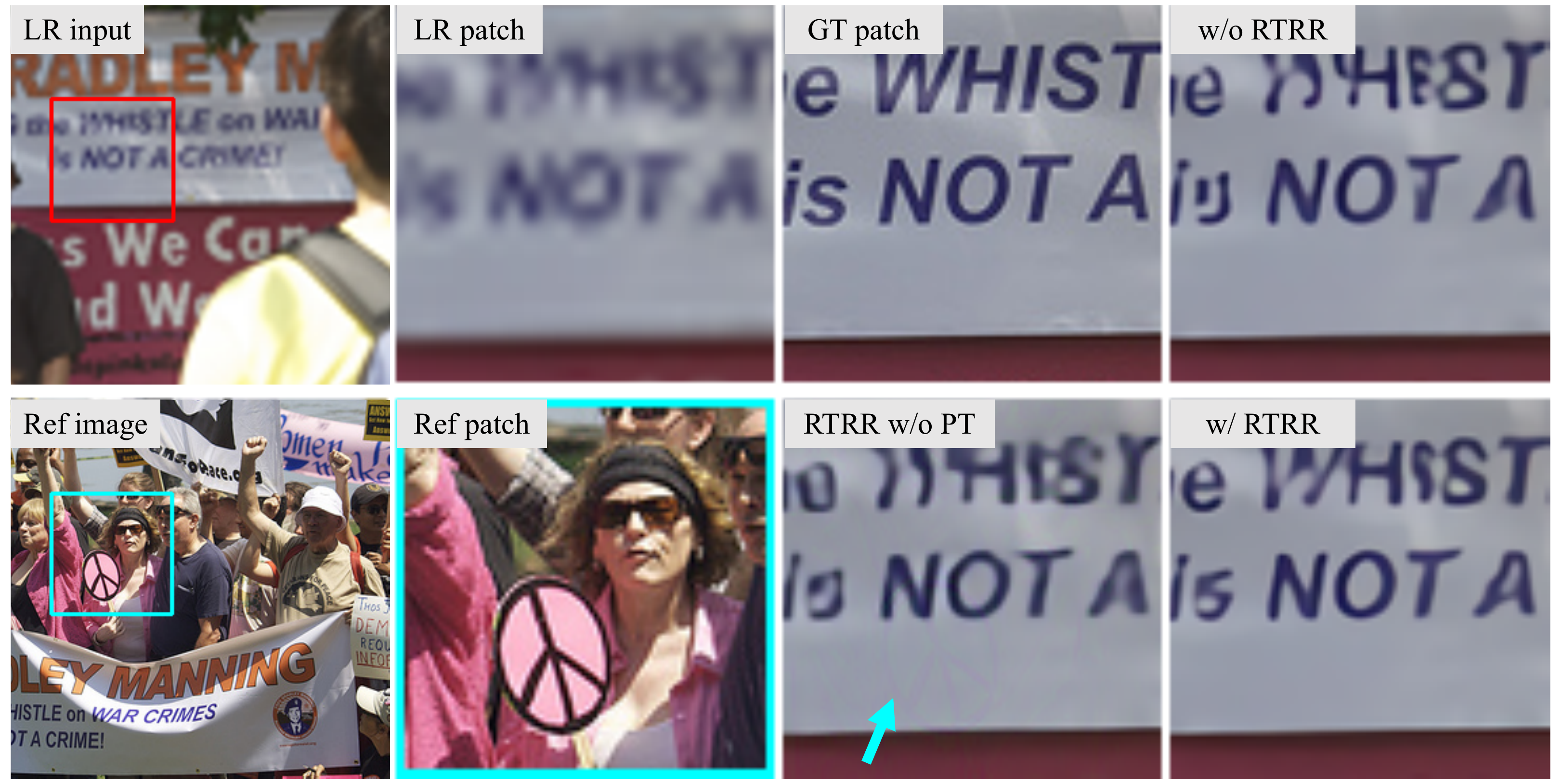}
	\end{center}
	\caption{Qualitative comparisons of ablation study on RTRR. The first column shows the input image and the reference image. On the right side of them, we zoom in a small region for better analysis. To take a close look, we find that a ghost of {\color[rgb]{0.0, 1.0, 1.0}\textit{handheld fan}} from reference image occurs in the SR result of a model with reciprocal learning but no perspective transformation (RTRR w/o PT).}
	\label{fig:ablation_rtrr}
\end{figure}

\begin{figure}[b]
	\begin{center}
		\includegraphics[width=0.99\columnwidth]{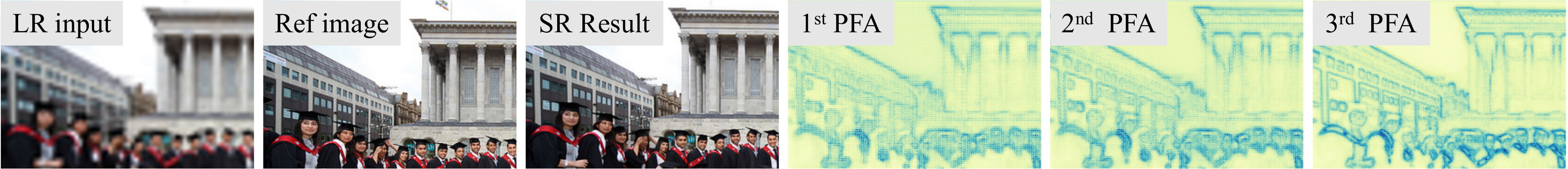}
	\end{center}
	\caption{Feature visualization at different PFA states on $4\times$ feature scale. The quality of features is improved progressively.}
	\label{fig:ablation_pfa}
\end{figure}

\begin{figure}[t]
    \centering
    \subfloat[]{\includegraphics[height=3.8cm, ]{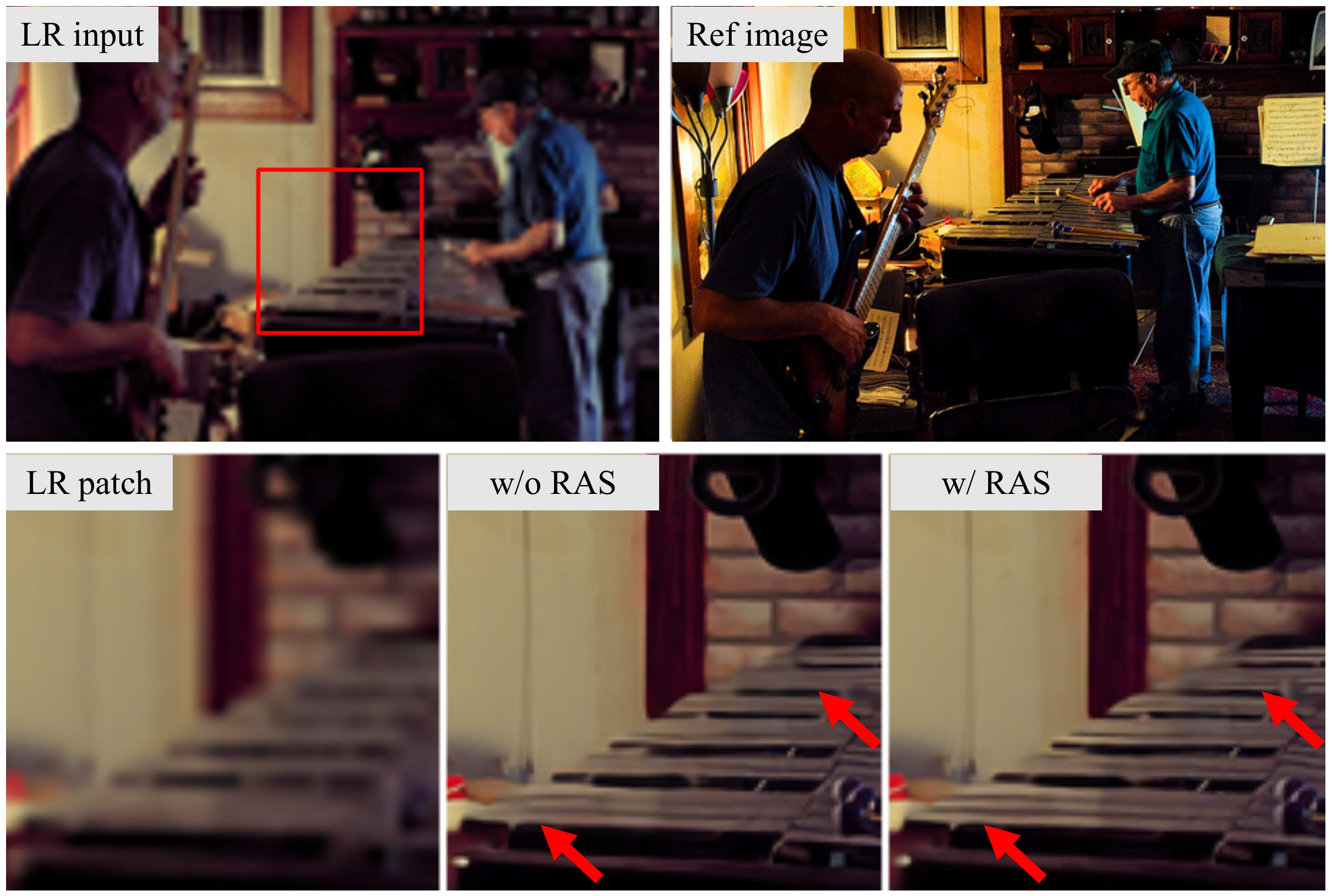}} \ 
    \subfloat[]{\includegraphics[height=3.8cm, ]{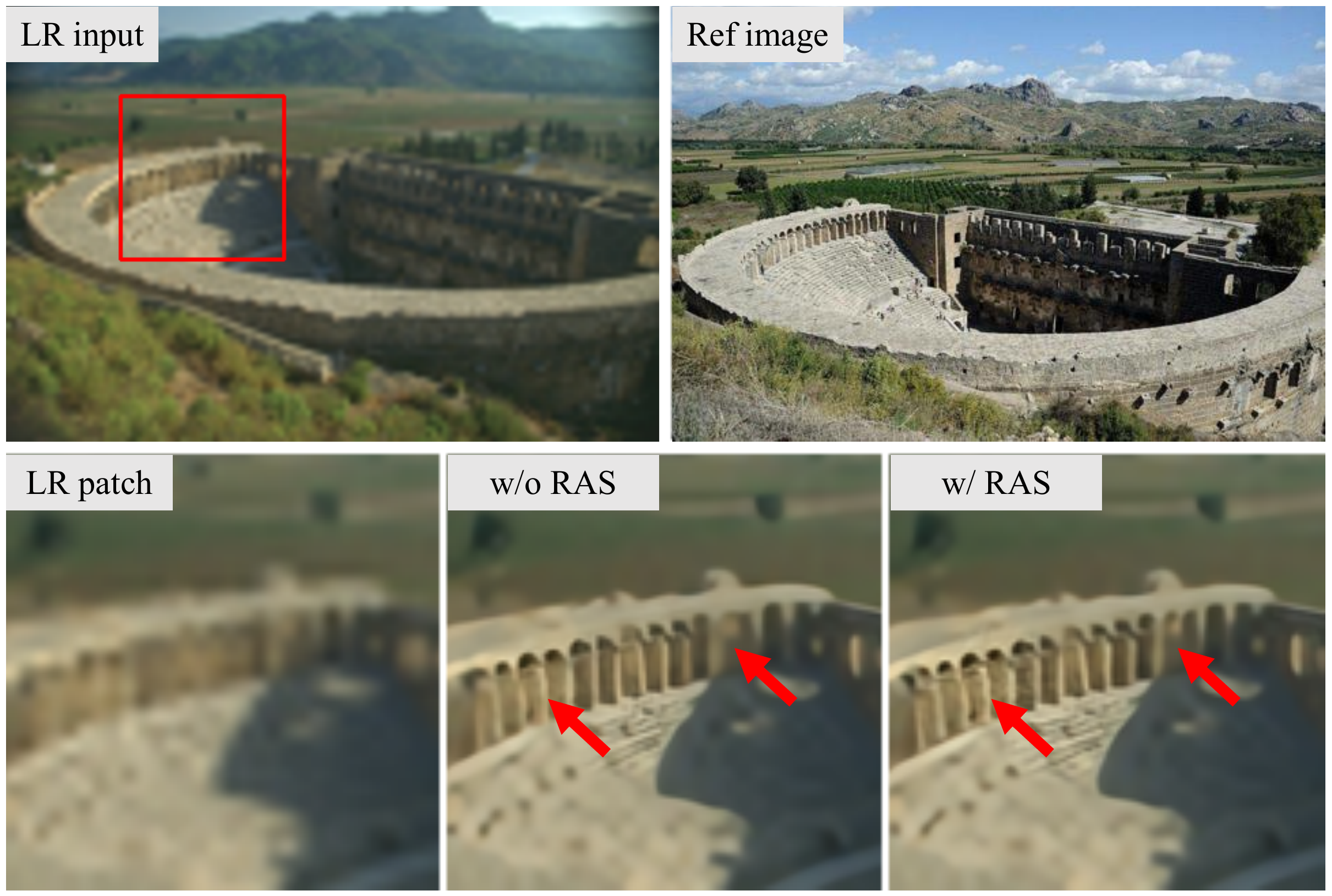}}
\caption{Qualitative comparisons of ablation study on reference-aware feature selection. There are color and style differences between input and reference images in (a) and (b). Models equipped with RAS can reproduce sharper textures.}
\label{fig:ablation_ras}
\end{figure}

\subsection{Ablation Study}
\label{sec:ablation}

In this section, we verify the effectiveness of different modules in our approach. The testing set of CUFED5 is used for evaluating model. As shown in left of the Table~\ref{table:ablations}, starting with a $C^2$-Matching baseline model, we separately evaluate the impact of the progressive feature alignment (PFA) and the reference-aware feature selection (RAS) in the proposed Progressive Feature Alignment and Selection (PFAS). Then we demonstrate the effectiveness of the RTRR training strategies.

\noindent\textbf{Reciprocal Target-Reference Reconstruction.} We introduce the RTRR training strategies on $C^2$-Matching~\cite{jiang2021robust}, MASA~\cite{lu2021masa} and TTSR~\cite{yang2020learning}, since these methods are open-sourced. The right part of the Table~\ref{table:ablations} shows the influence of the RTRR. It can be seen that all methods have consistent improvement, especially a 0.24 dB in $C^2$-Matching. We notice that the improvement of MASA and TTSR with RTRR is slightly smaller, because the SR results of these two methods are a little worse and RTRR relies on the SR results to do the second time RefSR.

We study the effect of the coefficient $\lambda_{RTRR}$ of the $\mathcal{L}_{RTRR}$ loss. As shown in Fig.~\ref{fig:user_loss} (b), The PSNR of the model with the RTRR is significantly higher than that without. But as the $\lambda_{RTRR}$ increases beyond 0.4, the results start to drop. It's easy to understand because in the inference phase, we use the first time RRSR instead of the second which with the SR image $X_{SR}$ as a reference image. There is a data distribution gap between the SR result and a real HR image when used as a reference, so $\lambda_{RTRR}$ should not be too large.

Fig.~\ref{fig:ablation_rtrr} intuitively illustrates the important role of the perspective transformation used to generate the input image $Y_{LR}^{\!\mathcal{P}}$ for second time RRSR. The SR result image of a model with reciprocal learning but no perspective transformation have reference image textures and the reason has been explained in Sec 3.1, these textures are used for generating SR result of the second time RRSR. From Fig.~\ref{fig:ablation_rtrr} we can also see with the RRTR, the output SR images have more clear and realistic textures. More experiments and discussions of perspective transformation are in supplementary material.

\noindent\textbf{Progressive Feature Alignment and Selection.}
As indicated in the left of the Table~\ref{table:ablations}, We verify the effects of the PFA which has an PSNR improvement of 0.23dB. Fig.~\ref{fig:ablation_pfa} shows the visualized feature maps after different alignments module on $4\times$ feature scale. It can be seen that the textures of feature maps gradually become clear after passing through multiple alignments.

Furthermore, we assess the impact of RAS and get a PSNR gain of 0.07dB. Fig.~\ref{fig:ablation_ras} shows the the influence of the RAS module. When the reference images are very different in color, lighting, style, etc., with the RAS module, our method can better select reference features to generate more clear textures. Finally, we analyz the influence of the number of learnable filters in RAS. We try the learnable filters' number of 8, 16 and 32, and find that compared to 16, the PSNRs of the others decrease slightly by around 0.02 dB. We also compare RAS with feature alignment methods based on statistics in supplementary material.

\FloatBarrier\section{Conclusion}

In this paper, we propose a novel reciprocal learning strategy for reference-based image super-resolution (RefSR). In addition, in order to transfer more accurate reference features, we design a progressive feature alignment and selection module. Extensive experimental results have demonstrated the superiority of our proposed RefSR model against recent state-of-the-art framework named $C^2$-Matching. We also validate that the reciprocal learning strategy is model-agnostic and it can be applied to improve arbitrary RefSR models. Specifically, our reciprocal learning method consistently improves three recent state-of-the-art RefSR frameworks. Combining the reciprocal learning and progressive feature alignment and selection strategies, we set new state-of-the-art RefSR performances on multiple benchmarks.
\\

\noindent \textbf{Acknowledgements} We thank Qing Chang, He Zheng, and  anonymous reviewers for helpful discussions. This work was supported in part by the Major Project for New Generation of AI (No.2018AAA0100400), the National Natural Science Foundation of China (No. 61836014, No. U21B2042, No. 62072457, No. 62006231) and in part by the Baidu Collaborative Research Project.

%
%
\bibliographystyle{splncs04}
\bibliography{7808}

\begin{thebibliography}{10}
\providecommand{\url}[1]{\texttt{#1}}
\providecommand{\urlprefix}{URL }
\providecommand{\doi}[1]{https://doi.org/#1}

\bibitem{dai2019second}
Dai, T., Cai, J., Zhang, Y., Xia, S.T., Zhang, L.: Second-order attention
  network for single image super-resolution. In: Proceedings of the IEEE
  Conference on Computer Vision and Pattern Recognition (CVPR) (2019)

\bibitem{dong2014learning}
Dong, C., Loy, C.C., He, K., Tang, X.: Learning a deep convolutional network
  for image super-resolution. In: Proceedings of the European Conference on
  Computer Vision (ECCV) (2014)

\bibitem{goodfellow2014gan}
Goodfellow, I., Pouget-Abadie, J., Mirza, M., Xu, B., Warde-Farley, D., Ozair,
  S., Courville, A., Bengio, Y.: Generative adversarial nets. In: Advances in
  Neural Information Processing Systems (NeurIPS) (2014)

\bibitem{greenspan2009super}
Greenspan, H.: Super-resolution in medical imaging. The computer journal
  (2009)

\bibitem{ha2016hypernetworks}
Ha, D., Dai, A., Le, Q.V.: Hypernetworks. In: International Conference on
  Learning Representations (ICLR) (2017)

\bibitem{he2016dual}
He, D., Xia, Y., Qin, T., Wang, L., Yu, N., Liu, T.Y., Ma, W.Y.: Dual learning
  for machine translation. In: Advances in Neural Information Processing
  Systems (NeurIPS) (2016)

\bibitem{he2016deep}
He, K., Zhang, X., Ren, S., Sun, J.: Deep residual learning for image
  recognition. In: Proceedings of the IEEE Conference on Computer Vision and
  Pattern Recognition (CVPR) (2016)

\bibitem{holden2011daostorm}
Holden, S.J., Uphoff, S., Kapanidis, A.N.: Daostorm: An algorithm for
  high-density super-resolution microscopy. Nature methods  (2011)

\bibitem{hu2018squeeze}
Hu, J., Shen, L., Sun, G.: Squeeze-and-excitation networks. In: Proceedings of
  the IEEE Conference on Computer Vision and Pattern Recognition (CVPR) (2018)

\bibitem{hu2019meta}
Hu, X., Mu, H., Zhang, X., Wang, Z., Tan, T., Sun, J.: Meta-sr: A
  magnification-arbitrary network for super-resolution. In: Proceedings of the
  IEEE Conference on Computer Vision and Pattern Recognition (CVPR) (2019)

\bibitem{huang2015single}
Huang, J.B., Singh, A., Ahuja, N.: Single image super-resolution from
  transformed self-exemplars. In: Proceedings of the IEEE Conference on
  Computer Vision and Pattern Recognition (CVPR) (2015)

\bibitem{ioffe2015batch}
Ioffe, S., Szegedy, C.: Batch normalization: Accelerating deep network training
  by reducing internal covariate shift. In: Proceedings of the International
  Conference on Machine Learning (ICML) (2015)

\bibitem{jia2016dynamic}
Jia, X., De~Brabandere, B., Tuytelaars, T., Gool, L.V.: Dynamic filter
  networks. In: Advances in Neural Information Processing Systems (NeurIPS)
  (2016)

\bibitem{jiang2021reciprocal}
Jiang, H., Xu, Y., Cheng, Z., Pu, S., Niu, Y., Ren, W., Wu, F., Tan, W.:
  Reciprocal feature learning via explicit and implicit tasks in scene text
  recognition. In: International Conference on Document Analysis and
  Recognition (ICDAR) (2021)

\bibitem{jiang2021robust}
Jiang, Y., Chan, K.C., Wang, X., Loy, C.C., Liu, Z.: Robust reference-based
  super-resolution via c2-matching. In: Proceedings of the IEEE Conference on
  Computer Vision and Pattern Recognition (CVPR) (2021)

\bibitem{johnson2016perceptual}
Johnson, J., Alahi, A., Fei-Fei, L.: Perceptual losses for real-time style
  transfer and super-resolution. In: Proceedings of the European Conference on
  Computer Vision (ECCV) (2016)

\bibitem{ledig2017photo}
Ledig, C., Theis, L., Husz{\'a}r, F., Caballero, J., Cunningham, A., Acosta,
  A., Aitken, A., Tejani, A., Totz, J., Wang, Z., Shi, W.: Photo-realistic
  single image super-resolution using a generative adversarial network. In:
  Proceedings of the IEEE Conference on Computer Vision and Pattern Recognition
  (CVPR) (2017)

\bibitem{lim2017enhanced}
Lim, B., Son, S., Kim, H., Nah, S., Lee, K.M.: Enhanced deep residual networks
  for single image super-resolution. In: Proceedings of the IEEE Conference on
  Computer Vision and Pattern Recognition (CVPR) Workshops (2017)

\bibitem{liu2018non}
Liu, D., Wen, B., Fan, Y., Loy, C.C., Huang, T.S.: Non-local recurrent network
  for image restoration. In: Advances in Neural Information Processing Systems
  (NeurIPS) (2018)

\bibitem{lu2021masa}
Lu, L., Li, W., Tao, X., Lu, J., Jia, J.: Masa-sr: Matching acceleration and
  spatial adaptation for reference-based image super-resolution. In:
  Proceedings of the IEEE Conference on Computer Vision and Pattern Recognition
  (CVPR) (2021)

\bibitem{ma2020weightnet}
Ma, N., Zhang, X., Huang, J., Sun, J.: Weightnet: Revisiting the design space
  of weight networks. In: Proceedings of the European Conference on Computer
  Vision (ECCV) (2020)

\bibitem{matsui2017sketch}
Matsui, Y., Ito, K., Aramaki, Y., Fujimoto, A., Ogawa, T., Yamasaki, T.,
  Aizawa, K.: Sketch-based manga retrieval using manga109 dataset. Multimedia
  Tools and Applications  (2017)

\bibitem{mei2021image}
Mei, Y., Fan, Y., Zhou, Y.: Image super-resolution with non-local sparse
  attention. In: Proceedings of the IEEE Conference on Computer Vision and
  Pattern Recognition (CVPR) (2021)

\bibitem{mei2020image}
Mei, Y., Fan, Y., Zhou, Y., Huang, L., Huang, T.S., Shi, H.: Image
  super-resolution with cross-scale non-local attention and exhaustive
  self-exemplars mining. In: Proceedings of the IEEE Conference on Computer
  Vision and Pattern Recognition (CVPR) (2020)

\bibitem{mildenhall2018burst}
Mildenhall, B., Barron, J.T., Chen, J., Sharlet, D., Ng, R., Carroll, R.: Burst
  denoising with kernel prediction networks. In: Proceedings of the IEEE
  Conference on Computer Vision and Pattern Recognition (CVPR) (2018)

\bibitem{pham2021meta}
Pham, H., Dai, Z., Xie, Q., Le, Q.V.: Meta pseudo labels. In: Proceedings of
  the IEEE Conference on Computer Vision and Pattern Recognition (CVPR) (2021)

\bibitem{sajjadi2017enhancenet}
Sajjadi, M.S., Scholkopf, B., Hirsch, M.: Enhancenet: Single image
  super-resolution through automated texture synthesis. In: Proceedings of the
  IEEE Conference on Computer Vision and Pattern Recognition (CVPR) (2017)

\bibitem{shim2020robust}
Shim, G., Park, J., Kweon, I.S.: Robust reference-based super-resolution with
  similarity-aware deformable convolution. In: Proceedings of the IEEE
  Conference on Computer Vision and Pattern Recognition (CVPR) (2020)

\bibitem{simonyan2014vggnet}
Simonyan, K., Zisserman, A.: Very deep convolutional networks for large-scale
  image recognition. In: International Conference on Learning Representations
  (ICLR) (2015)

\bibitem{sitzmann2019scene}
Sitzmann, V., Zollh{\"o}fer, M., Wetzstein, G.: Scene representation networks:
  Continuous 3d-structure-aware neural scene representations. In: Advances in
  Neural Information Processing Systems (NeurIPS) (2019)

\bibitem{sun2020reciprocal}
Sun, H., Zhao, Z., He, Z.: Reciprocal learning networks for human trajectory
  prediction. In: Proceedings of the IEEE Conference on Computer Vision and
  Pattern Recognition (CVPR) (2020)

\bibitem{sun2012super}
Sun, L., Hays, J.: Super-resolution from internet-scale scene matching. In:
  IEEE International Conference on Computational Photography (ICCP) (2012)

\bibitem{tian2020conditional}
Tian, Z., Shen, C., Chen, H.: Conditional convolutions for instance
  segmentation. In: Proceedings of the European Conference on Computer Vision
  (ECCV) (2020)

\bibitem{wang2021learning}
Wang, L., Wang, Y., Lin, Z., Yang, J., An, W., Guo, Y.: Learning a single
  network for scale-arbitrary super-resolution. In: Proceedings of the IEEE
  International Conference on Computer Vision (ICCV) (2021)

\bibitem{wang2021dual}
Wang, T., Xie, J., Sun, W., Yan, Q., Chen, Q.: Dual-camera super-resolution
  with aligned attention modules. In: Proceedings of the IEEE International
  Conference on Computer Vision (ICCV) (2021)

\bibitem{wang2018esrgan}
Wang, X., Yu, K., Wu, S., Gu, J., Liu, Y., Dong, C., Qiao, Y., Loy, C.C.:
  Esrgan: Enhanced super-resolution generative adversarial networks. In:
  Proceedings of the European Conference on Computer Vision (ECCV) Workshops
  (2018)

\bibitem{wang2016event}
Wang, Y., Lin, Z., Shen, X., Mech, R., Miller, G., Cottrell, G.W.:
  Event-specific image importance. In: Proceedings of the IEEE Conference on
  Computer Vision and Pattern Recognition (CVPR) (2016)

\bibitem{xia2020basis}
Xia, Z., Perazzi, F., Gharbi, M., Sunkavalli, K., Chakrabarti, A.: Basis
  prediction networks for effective burst denoising with large kernels. In:
  Proceedings of the IEEE Conference on Computer Vision and Pattern Recognition
  (CVPR) (2020)

\bibitem{yang2019condconv}
Yang, B., Bender, G., Le, Q.V., Ngiam, J.: Condconv: Conditionally
  parameterized convolutions for efficient inference. In: Advances in Neural
  Information Processing Systems (NeurIPS) (2019)

\bibitem{yang2020learning}
Yang, F., Yang, H., Fu, J., Lu, H., Guo, B.: Learning texture transformer
  network for image super-resolution. In: Proceedings of the IEEE Conference on
  Computer Vision and Pattern Recognition (CVPR) (2020)

\bibitem{yi2017dualgan}
Yi, Z., Zhang, H., Tan, P., Gong, M.: Dualgan: Unsupervised dual learning for
  image-to-image translation. In: Proceedings of the IEEE International
  Conference on Computer Vision (ICCV) (2017)

\bibitem{yue2013landmark}
Yue, H., Sun, X., Yang, J., Wu, F.: Landmark image super-resolution by
  retrieving web images. IEEE Transactions on Image Processing (TIP)  (2013)

\bibitem{zagalsky2021design}
Zagalsky, A., Te'eni, D., Yahav, I., Schwartz, D.G., Silverman, G., Cohen, D.,
  Mann, Y., Lewinsky, D.: The design of reciprocal learning between human and
  artificial intelligence. Proceedings of the ACM on Human-Computer Interaction
   (2021)

\bibitem{zhang2010super}
Zhang, L., Zhang, H., Shen, H., Li, P.: A super-resolution reconstruction
  algorithm for surveillance images. Signal Processing  (2010)

\bibitem{zhang2019ranksrgan}
Zhang, W., Liu, Y., Dong, C., Qiao, Y.: Ranksrgan: Generative adversarial
  networks with ranker for image super-resolution. In: Proceedings of the IEEE
  International Conference on Computer Vision (ICCV) (2019)

\bibitem{zhang2018image}
Zhang, Y., Li, K., Li, K., Wang, L., Zhong, B., Fu, Y.: Image super-resolution
  using very deep residual channel attention networks. In: Proceedings of the
  European Conference on Computer Vision (ECCV) (2018)

\bibitem{zhang2019residual}
Zhang, Y., Li, K., Li, K., Zhong, B., Fu, Y.: Residual non-local attention
  networks for image restoration. In: International Conference on Learning
  Representations (ICLR) (2019)

\bibitem{zhang2019image}
Zhang, Z., Wang, Z., Lin, Z., Qi, H.: Image super-resolution by neural texture
  transfer. In: Proceedings of the IEEE Conference on Computer Vision and
  Pattern Recognition (CVPR) (2019)

\bibitem{zheng2018crossnet}
Zheng, H., Ji, M., Wang, H., Liu, Y., Fang, L.: Crossnet: An end-to-end
  reference-based super resolution network using cross-scale warping. In:
  Proceedings of the European Conference on Computer Vision (ECCV) (2018)

\end{thebibliography}
\end{document}